\documentclass{article}
\usepackage{ijcai19}

\usepackage{times}
\usepackage{soul}
\usepackage{url}
\usepackage[hidelinks]{hyperref}
\usepackage[utf8]{inputenc}
\usepackage[small]{caption}
\usepackage{graphicx}
\usepackage{amsmath}
\usepackage{booktabs}
\usepackage{amssymb}
\usepackage{algorithmic,algorithm}
\usepackage{subcaption}

\usepackage{bm}

\newcommand{\paratitle}[1]{\vspace{1.5ex}\noindent \textbf{#1\quad}}

\title{Experience Replay Optimization}

\author{
Daochen Zha \and
Kwei-Herng Lai \and
Kaixiong Zhou \And
Xia Hu
\affiliations
Department of Computer Science and Engineering, Texas A\&M University
\emails
\{daochen.zha, khlai037, zkxiong, xiahu\}@tamu.edu
}

\begin{document}

\maketitle

\begin{abstract}

Experience replay enables reinforcement learning agents to memorize and reuse past experiences, just as humans replay memories for the situation at hand. Contemporary off-policy algorithms either replay past experiences uniformly or utilize a rule-based replay strategy, which may be sub-optimal. In this work, we consider \emph{learning} a replay policy to optimize the cumulative reward. Replay learning is challenging because the replay memory is noisy and large, and the cumulative reward is unstable. To address these issues, we propose a novel experience replay optimization~(ERO) framework which alternately updates two policies: the agent policy, and the replay policy. The agent is updated to maximize the cumulative reward based on the replayed data, while the replay policy is updated to provide the agent with the most useful experiences. The conducted experiments on various continuous control tasks demonstrate the effectiveness of ERO, empirically showing promise in experience replay learning to improve the performance of off-policy reinforcement learning algorithms.
\end{abstract}

\section{Introduction}

Experience replay mechanism~\cite{lin1992self,lin1993reinforcement} plays a significant role in deep reinforcement learning, enabling the agent to memorize and reuse past experiences. It is demonstrated that experience replay greatly stabilizes the training process and improves the sample efficiency by breaking the temporal correlations~\cite{mnih2013playing,mnih2015human,wang2016sample,van2016deep,andrychowicz2017hindsight,pan2018organizing,sutton2018reinforcement}. Current off-policy algorithms usually adopt a uniform sampling strategy which replays past experiences with equal frequency. However, the uniform sampling cannot reflect the different importance of past experiences: the agent can usually learn more efficiently from some experiences than from others, just as humans tend to replay crucial experiences and generalize them to the situation at hand~\cite{shohamy2015integrating}.

Recently, some rule-based replay strategies have been studied to prioritize important experiences. One approach is to directly prioritize the transitions\footnote{Transition and experience are considered exchangeable in this work when the context has no ambiguity.} with higher temporal difference~(TD) errors~\cite{schaul2016prioritized}. This simple replay rule is shown to improve the performance of deep Q-network on Atari environments and is demonstrated to be a useful ingredient in Rainbow~\cite{hessel2018rainbow}. Other studies indirectly prioritize experiences through managing the replay memory. Adopting different sizes of the replay memory~\cite{zhang2017deeper,liu2017effects} and selectively remembering/forgetting some experiences with simple rules~\cite{novati2018remember} are shown to affect the performance greatly. However, the rule-based replay strategies may be sub-optimal and may not be able to adapt to different tasks or reinforcement learning algorithms. We are thus interested in studying how we can optimize the replay strategy towards more efficient use of the replay memory.

In the neuroscience domain, the recently proposed normative theory for memory access suggests that a rational agent ought to replay the memories that lead to the most rewarding future decisions~\cite{mattar2018prioritized}. For instance, when a human being learns to run, she tends to utilize the memories that can mostly accelerate the learning process; in this context, the memories could relate to walking. We are thus motivated to use the feedback from the environment as a rewarding signal to adjust the replay strategy. Specifically, apart from the agent policy, we consider learning an additional replay policy, aiming at sampling the most useful experiences for the agent to maximize the cumulative reward. A learning-based replay policy is promising because it can potentially find more useful experiences to train the agent and may better adapt to different tasks and algorithms.

However, it is nontrivial to model the replay policy for several challenges. First, transitions in the memory are quite noisy due to the randomness of the environment. Second, the replay memory is typically large. For example, the common memory size for the benchmark off-policy algorithm deep deterministic policy gradient~(DDPG)~\cite{lillicrap2016continuous} can reach $10^6$. The replay policy needs to properly filter out the most useful ones among all the transitions in the memory.  Third, the cumulative reward is unstable, also due to the environmental randomness. Therefore, it is challenging to effectively and efficiently learn the replay policy.

To address the above issues, in this paper, we propose experience replay optimization~(ERO) framework, which alternately updates two policies: the agent policy, and the replay policy. Specifically, we investigate how to efficiently replay the most useful experiences from the replay buffer, and how we can make use of the environmental feedback to update the replay policy. The main contributions of this work are summarized as follows:

\begin{itemize}
  \item Formulate experience replay as a learning problem.
  \item Propose ERO, a general framework for effective and efficient use of replay memory. A priority vector is maintained to sample subsets of transitions for efficient replaying. The replay policy is updated based on the improvement of cumulative reward.
  \item Develop an instance of ERO by applying it to the benchmark off-policy algorithm DDPG.
  \item Conduct experiments on $8$ continuous control tasks from OpenAI Gym to evaluate our framework. Empirical results show promise in experience replay learning to improve the performance of off-policy algorithms.
\end{itemize}

\section{Problem Statement}
We consider standard reinforcement learning~(RL) which is represented by a sextuple $(\mathcal{S}, \mathcal{A}, \mathcal{P}_T, \mathcal{R}, \gamma, p_0)$, where $\mathcal{S}$ is the set of states, $\mathcal{A}$ is the set of actions, $\mathcal{P}_T: \mathcal{S} \times \mathcal{A} \to \mathcal{S}$ is the state transition function, $\mathcal{R}: \mathcal{S} \times \mathcal{A} \times \mathcal{S} \to \mathbb{R}$ is the reward function, $\gamma \in (0, 1)$ is the discount factor, and $p_0$ is the distribution of the initial state. At each timestep $t$, an agent takes action $a_t \in \mathcal{A}$ in state $s_t \in \mathcal{S}$ and observes the next state $s_{t+1}$ with a scalar reward $r_t$, which results in a quadruple $(s_t, a_t, r_t, s_{t+1})$, also called \emph{transition}. This also leads to a \emph{trajectory} of states, actions and rewards $(s_1, a_1, r_1, s_2, a_2, r_2, ...)$. The objective is to learn a policy $\pi: \mathcal{S} \to \mathcal{A}$ that maximizes the cumulative reward $R = \mathbb{E}[\sum_{t=1}^{\infty}\gamma^t r_t]$. An off-policy agent makes use of an experience replay buffer, denoted as $\mathcal{B}$. At each timestep $t$, the agent interacts with the environment and stores transition $(s_t, a_t, r_t, s_{t+1})$ into $\mathcal{B}$. Let $\mathcal{B}_i$ denote the transition in $\mathcal{B}$ at position $i$. Then for each training step, the agent is updated by using a batch of transitions $\{\mathcal{B}_i\}$ sampled from the buffer.

Based on the notations defined above, we formulate the problem of learning a replay policy as follows. Given a task $\mathcal{T}$, an off-policy agent $\Lambda$ and the experience replay buffer $\mathcal{B}$, we aim at learning a replay policy $\phi$ which at each training step samples a batch of transitions $\{\mathcal{B}_i\}$ from $\mathcal{B}$ to train agent $\Lambda$, i.e., learning the mapping $\phi:\mathcal{B} \to \{\mathcal{B}_i\}$, such that better performance can be achieved by $\Lambda$ on task $\mathcal{T}$ in terms of cumulative reward and efficiency.

\begin{figure}[t]
\centering
\includegraphics[width=8.5cm]{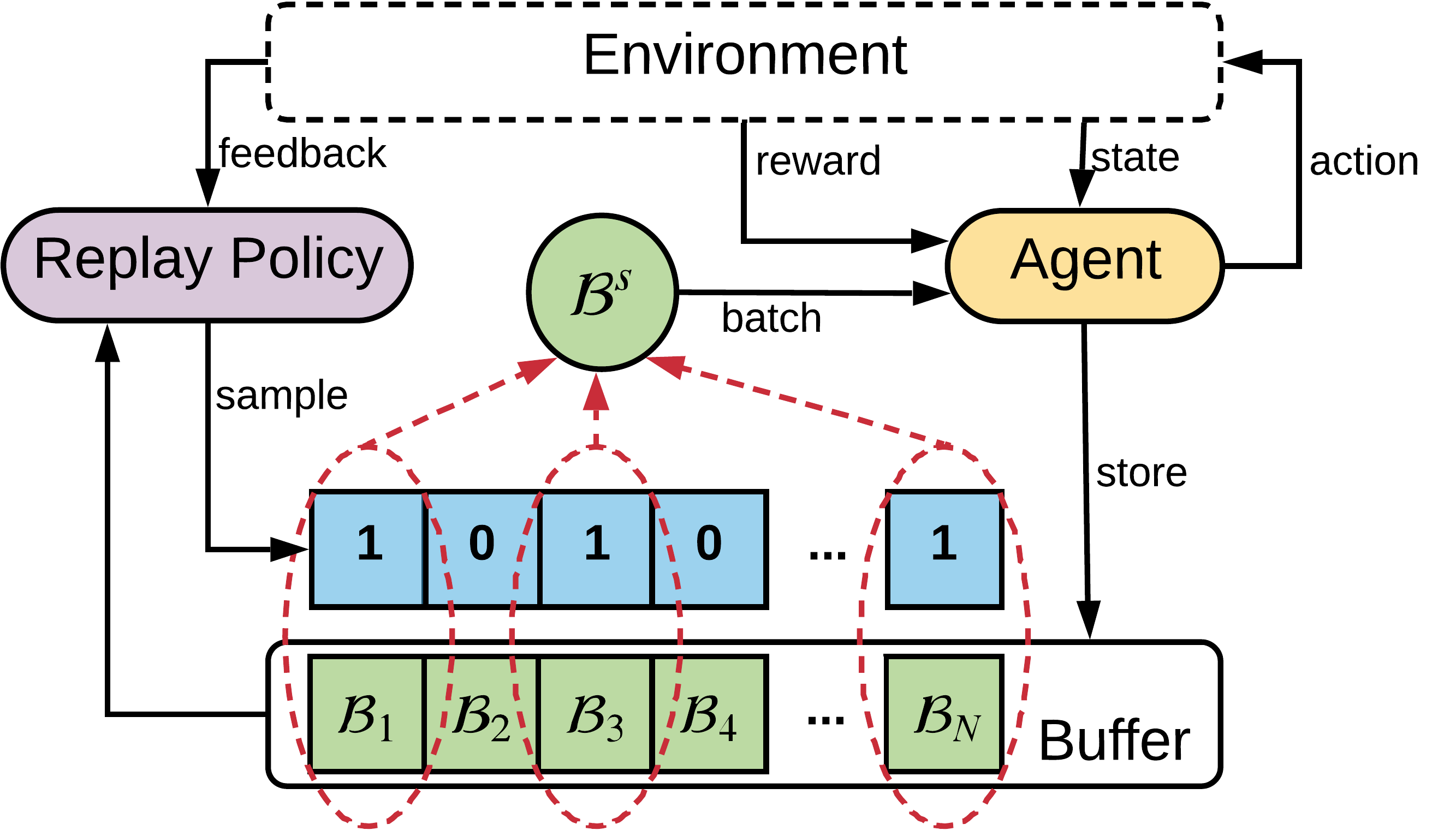}
\caption{An overview of experience replay optimization (ERO). The reinforcement learning agent interacts with the environment and stores the transition into the buffer. In training, the replay policy generates a vector to sample a subset of transitions $\mathcal{B}^s$ from the buffer, where $1$ indicates that the corresponding transition is selected. The sampled transitions are then used to update the agent.}
\label{label:fig1}
\end{figure}

\section{Methodology}
 
Figure~\ref{label:fig1} shows an overview of the ERO framework. The core idea of the proposed framework is to use a replay policy to sample a subset of transitions from the buffer, for updating the reinforcement learning agent. Here, the replay policy generates a 0-1 boolean vector to guide subset sampling, where $1$ indicates that the corresponding transition is selected~(detailed in Section~\ref{label:sec3-1}). The replay policy indirectly teaches the agent by controlling what historical transitions are used, and is adjusted according to the feedback which is defined as the improvement of the performance~(detailed in Section~\ref{label:sec3-2}). Our framework is general and can be applied to contemporary off-policy algorithms. Next, we elaborate the details of the proposed ERO framework.

\subsection{Sampling with Replay Policy}
\label{label:sec3-1}

In this subsection, we formulate a learning-based replay policy to sample transitions from the buffer.

Let $\mathcal{B}_i$ be a transition in buffer $\mathcal{B}$ with a associated feature vector $\textbf{f}_{\mathcal{B}_i}$~(we empirically select some features, detailed in Section \ref{label:sec_exp}), where $i$ denotes the index. In ERO, the replay policy is described as a priority score function $\phi(\textbf{f}_{\mathcal{B}_i}|\theta^\phi) \in (0, 1)$, in which higher value indicates higher probability of a transition being replayed, $\phi$ denotes a function approximator which is a deep neural network, and $\theta^\phi$ are the corresponding parameters. For each $\mathcal{B}_i$, a score is calculated by the replay policy $\phi$. We further maintain a vector $\bm{\lambda}$ to store these scores:
\begin{equation}
    \bm{\lambda} = \{\phi(\textbf{f}_{\mathcal{B}_i}|\theta^\phi) | \mathcal{B}_i \in \mathcal{B} \} \in \mathbb{R}^{N},
\end{equation}
where $N$ is the number of transitions in $\mathcal{B}$, element $\bm{\lambda}_i$ is the priority score of the corresponding transition $\mathcal{B}_i$. Note that it is infeasible to keep all the priority scores up-to-date because the buffer size is usually large. To avoid expensive sweeps over the entire buffer, we adopt an updating strategy similar to~\cite{schaul2016prioritized}. Specifically, a score is updated only when the corresponding transition is replayed. With this approximation, some transitions with very low scores could remain almost never sampled. However, we have to consider the efficiency issue when developing replay policy. In our preliminary experiments, we find this approximation works well and significantly accelerates the sampling. Given the scores vector $\bm{\lambda}$, we then sample $\mathcal{B}^{s}$ according to 
\begin{align}
\begin{split}
\label{label:eqn2}
    \textbf{I} & \sim \mathrm{Bernoulli}(\bm{\lambda}), \\
    \mathcal{B}^{s} & = \{\mathcal{B}_i | \mathcal{B} _i \in \mathcal{B}\wedge \textbf{I}_i = 1 \},
\end{split}
\end{align}
where $\mathrm{Bernoulli}(\cdot)$ denotes the Bernoulli distribution, $\textbf{I}$ is an N-dimensional vector with element $\textbf{I}_i \in \{0, 1\}$. That is, $\mathcal{B}^{s}$ is the subset of $\mathcal{B}$ such that the transition is selected when the corresponding value in \textbf{I} is $1$.
In this sense, if transition $\mathcal{B}_{i}$ has higher priority, it will be more likely to have $\textbf{I}_i = 1$, hence being more likely to be in the subset $\mathcal{B}^s$. Then $\mathcal{B}^s$ is used to update the agent with standard procedures, i.e., mini-batch updates with uniform sampling. The binary vector \textbf{I}, which serves as a ``mask" to narrow down all the transitions to a smaller subset of transitions, indirectly affects the relaying.

\begin{algorithm}[t]
\caption{ERO enhanced DDPG}
\label{label:alg1}
\begin{algorithmic}[1]
\STATE Initialize policy $\pi$, replay policy $\phi$, buffer $\mathcal{B}$
\FOR{each iteration}
    \FOR{each timestep $t$}
        \STATE Select action $a_t$ according to $\pi$ and state $s_t$
        \STATE Execute action $a_t$ and observe $s_{t+1}$ and $r_t$
        \STATE Store transition $(s_t, a_t, r_t, s_{t+1})$ into $\mathcal{B}$
        \IF{episode is done}
            \STATE Calculate the cumulative reward $r^{c}_{\pi}$
            \IF{$r^{c}_{\pi'}$ $\neq$ null}
                \STATE $\mathcal{B}^s$ = UpdateReplayPolicy($r^{c}_{\pi}$, $r^{c}_{\pi'}$, $\mathcal{B}$)
            \ENDIF
            \STATE Set $r^{c}_{\pi'} \leftarrow r^{c}_{\pi}$
        \ENDIF
    \ENDFOR
    \FOR{each training step}
        \STATE Uniformly sample a batch $\{\mathcal{B}^s_i\}$ from $\mathcal{B}^s$
        \STATE Update the critic of $\pi$ with Eq. (\ref{label:eqn9}) and (\ref{label:eqn10})
        \STATE Update the actor of $\pi$ with Eq. (\ref{label:eqn11})
        \STATE Update target networks with Eq. (\ref{label:eqn12}) and (\ref{label:eqn13})
        \STATE Update ${\lambda}$ for each transition in $\{\mathcal{B}_i\}$ 
    \ENDFOR
\ENDFOR
\end{algorithmic}
\end{algorithm}
\subsection{Training with Policy Gradient}
\label{label:sec3-2}
This subsection describes how the replay policy is updated in ERO framework.

From the perspective of reinforcement learning, the binary vector \textbf{I} can be regarded as the ``action" taken by the replay policy. The replay-reward is defined as
\begin{equation}
\label{label:eqn3}
    r^{r} = r^{c}_{\pi} - r^{c}_{\pi'},
\end{equation}
where $r^{c}_{\pi'}$ and $r^{c}_{\pi}$ are the cumulative reward of the previous agent policy $\pi'$ and the agent policy $\pi$ updated based on \textbf{I}, respectively. The cumulative reward of $\pi$ is estimated by the recent episodes it performs\footnote{In our implementation, a replay-reward is computed and used to update the replay policy when one episode is finished. $r^{c}_{\pi}$ is estimated by the mean of recent $100$ episodes.}. The replay-reward can be interpreted as how much the action \textbf{I} helps the learning of the agent. The objective of the replay policy is to maximize the improvement:
\begin{equation}
    \mathcal{J} = \mathbb{E}_{\textbf{I}}[r^{r}].
\end{equation}
By using the REINFORCE trick~\cite{williams1992simple}, we can calculate the gradient of $\mathcal{J}$ w.r.t $\theta^\phi$:
\begin{align}
\begin{split}
    \bigtriangledown_{\theta^\phi} \mathcal{J} & = \bigtriangledown_{\theta^\phi} \mathbb{E}_{\textbf{I}}[r^{r}] \\
    & = \mathbb{E}_{\textbf{I}}[r^{r} \bigtriangledown_{\theta^\phi} \log \mathcal{P}(\textbf{I}|\phi)],
\end{split}
\end{align}
where $\mathcal{P}(\textbf{I}|\phi)$ is the probability of generating binary vector \textbf{I} given $\phi$, and $\phi$ is the abbreviation for $\phi(\textbf{f}_{\mathcal{B}_i}|\theta^\phi)$. Since each entry of \textbf{I} is sampled independently based on $\bm{\lambda}$, the term $\log \mathcal{P}(\textbf{I}|\phi)$ can be factorized as:
\begin{align}
\begin{split}
    \log \mathcal{P}(\textbf{I}|\phi) & =\sum_{i=1}^{N} \log \mathcal{P}(\textbf{I}_i|\phi) \\
    & = \sum_{i=1}^{N} [\textbf{I}_i \log \phi + (1-\textbf{I}_i) \log (1-\phi)],
\end{split}
\end{align}
where $N$ is the number of the transitions in $\mathcal{B}$, and $\textbf{I}_i$ is the binary selector for transition $\mathcal{B}_i$. The resulting policy gradient can be written as:
\begin{equation}
\label{label:eqn7}
      \bigtriangledown_{\theta^\phi} \mathcal{J} \approx r^{r} \sum_{i=1}^{N} \bigtriangledown_{\theta^\phi} [\textbf{I}_i \log \phi + (1-\textbf{I}_i) \log (1-\phi)].
\end{equation}
If we regard \textbf{I} as ``labels", then Eq.~(\ref{label:eqn7}) can be viewed as cross-entropy loss of the replay policy $\phi$, scaled by $r^{r}$. Intuitively, a current positive (negative) reward will encourage (discourage) the next action to be similar to the current action \textbf{I}. Thus, the replay policy is updated to maximize the replay-reward. At each replay updating step, the gradient of Eq.~(\ref{label:eqn7}) can be approximated by sub-sample a mini-batch of transitions from $\mathcal{B}$ to efficiently update the replay policy:

\begin{equation}
\label{label:eqn8}
       \sum_{j : \mathcal{B}_j \in \mathcal{B}^{batch}} r^{r} \bigtriangledown_{\theta^\phi} [\textbf{I}_j \log \phi + (1-\textbf{I}_j) \log (1-\phi)],
\end{equation}
where $\mathcal{B}^{batch}$ denotes a mini-batch of transitions sampled from $\mathcal{B}$. Note that the replay policy is updated only at the end of each episode. When an episode is finished, the replay-reward for the current action \textbf{I} is used to update the replay policy. The updates of the replay policy rely on the cumulative rewards of the recent episodes in the training process, but do not require generating new episodes.

\begin{algorithm}[t]
\caption{UpdateReplayPolicy}
\label{label:alg2}
\textbf{Input:}\\
\hspace*{\algorithmicindent}Cumulative reward of current policy $r^{c}_{\pi}$\\
\hspace*{\algorithmicindent}Cumulative reward of previous policy $r^{c}_{\pi'}$\\
\hspace*{\algorithmicindent}Buffer $\mathcal{B}$ \\
\textbf{Output:} \\
\hspace*{\algorithmicindent}Sampled subset $\mathcal{B}^s$
 \
\begin{algorithmic}[1]
    
    \STATE Calculate replay-reward based on Eq. (\ref{label:eqn3})
    \FOR{each replay updating step}
        \STATE Randomly sample a batch $\{\mathcal{B}_i\}$ from $\mathcal{B}$
        \STATE Update replay policy based on Eq. (\ref{label:eqn8})
    \ENDFOR
    \STATE Sample a subset $\mathcal{B}^s$ from $\mathcal{B}$ using Eq. (\ref{label:eqn2})
\end{algorithmic}
\end{algorithm}

\section{Application to Off-policy Algorithms}
In this section, we use the benchmark off-policy algorithm DDPG~\cite{lillicrap2016continuous} as an example to show how ERO is applied. Note that ERO could also be applied to other off-policy algorithms by using similar procedures.

DDPG is a model-free actor-critic algorithm consisting of an actor function $\mu(s|\theta^\mu)$ that specifies the current policy and a critic function $Q(s_t, a_t | \theta^Q)$. Here, $\theta^\mu$ and $\theta^Q$ are approximated by deep neural networks. In training, DDPG optimizes $\theta^Q$ by minimizing the following loss w.r.t $\theta^Q$:
\begin{equation}
\label{label:eqn9}
    \mathcal{L}(\theta^Q) = \frac{1}{N} \sum_t(y_t-Q(s_t,a_t|\theta^Q))^2,
\end{equation}
where
\begin{equation}
\label{label:eqn10}
    y_t = r(s_t,a_t) + \gamma Q(s_{t+1}, \mu(s_{t+1}|\theta^\mu)|\theta^Q).
\end{equation}
The actor network can be updated by applying chain rules to $\mathcal{J}$, the expected return over the state distribution, w.r.t $\theta^\mu$~\cite{silver2014deterministic}:

\begin{align}
\label{label:eqn11}
\begin{split}
    \bigtriangledown_{\theta^\mu} \mathcal{J} & \approx \mathbb{E}_{s_t} [\bigtriangledown_{\theta^\mu} Q(s_t,\mu(s_t|\theta^\mu)|\theta^Q)] \\
    & = \mathbb{E}_{s_t} [\bigtriangledown_\mu Q(s_t,\mu(s_t|\theta_\mu)|\theta^Q) \bigtriangledown_{\theta_\mu} \mu(s_t|\theta^\mu)].
\end{split}
\end{align}
To learn the non-linear function approximators in a stable manner, DDPG uses two corresponding target networks to update slowly:
\begin{align}
    \label{label:eqn12}
    \theta^{Q'} \leftarrow \tau \theta^Q + (1-\tau) \theta^{Q'}, \\
    \label{label:eqn13}
    \theta^{\mu'} \leftarrow \tau \theta^\mu + (1-\tau) \theta^{\mu'},
\end{align}
where $\tau \in (0,1]$, $Q'$ and $\mu'$ are the target critic network and target actor network, respectively. DDPG maintains an experience replay buffer. For each training step, the critic and the actor are updated by using a randomly sampled batch of transitions from the buffer.

To apply ERO framework to DDPG, we first sample a subset of the transitions from the buffer according to the replay policy, and then feed this subset of transitions to DDPG algorithm. The ERO enhanced DDPG is summarized in Algorithm~\ref{label:alg1}, where Line 20 updates the transition priority scores with the replay policy, and Algorithm~\ref{label:alg2} serves for replay learning purpose.

\begin{figure*}
  \centering
  \begin{subfigure}[b]{0.60\textwidth}
    \includegraphics[width=1.0\textwidth]{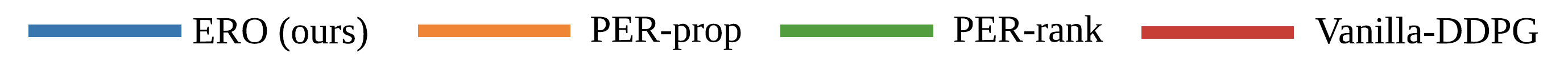}
  \end{subfigure}
  \vspace{-15pt}
  \begin{subfigure}[b]{0.25\textwidth}
    \includegraphics[width=\textwidth]{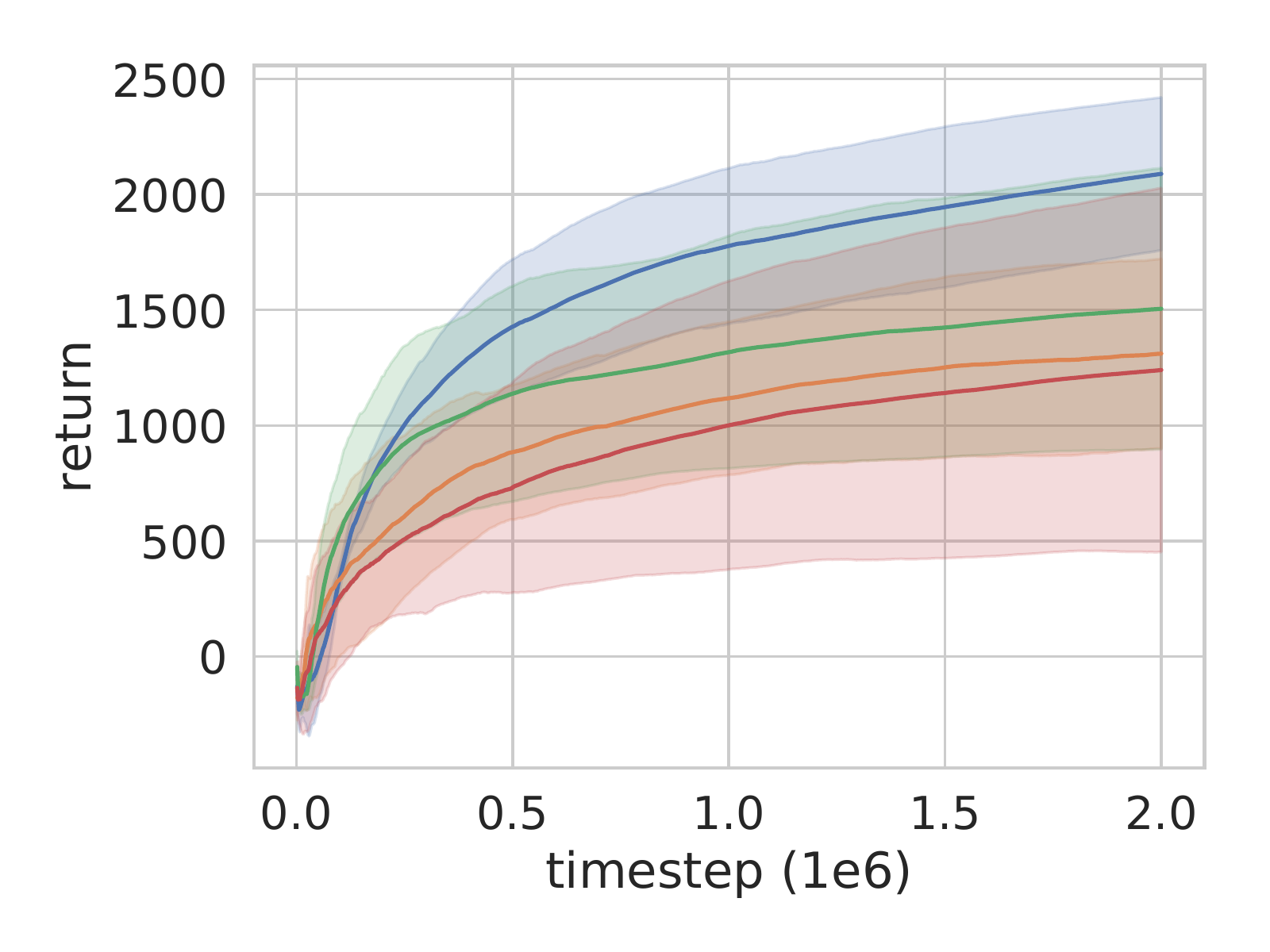}
    \caption{HalfCheetah}
    \vspace{15pt}
  \end{subfigure}%
  \begin{subfigure}[b]{0.25\textwidth}
    \includegraphics[width=\textwidth]{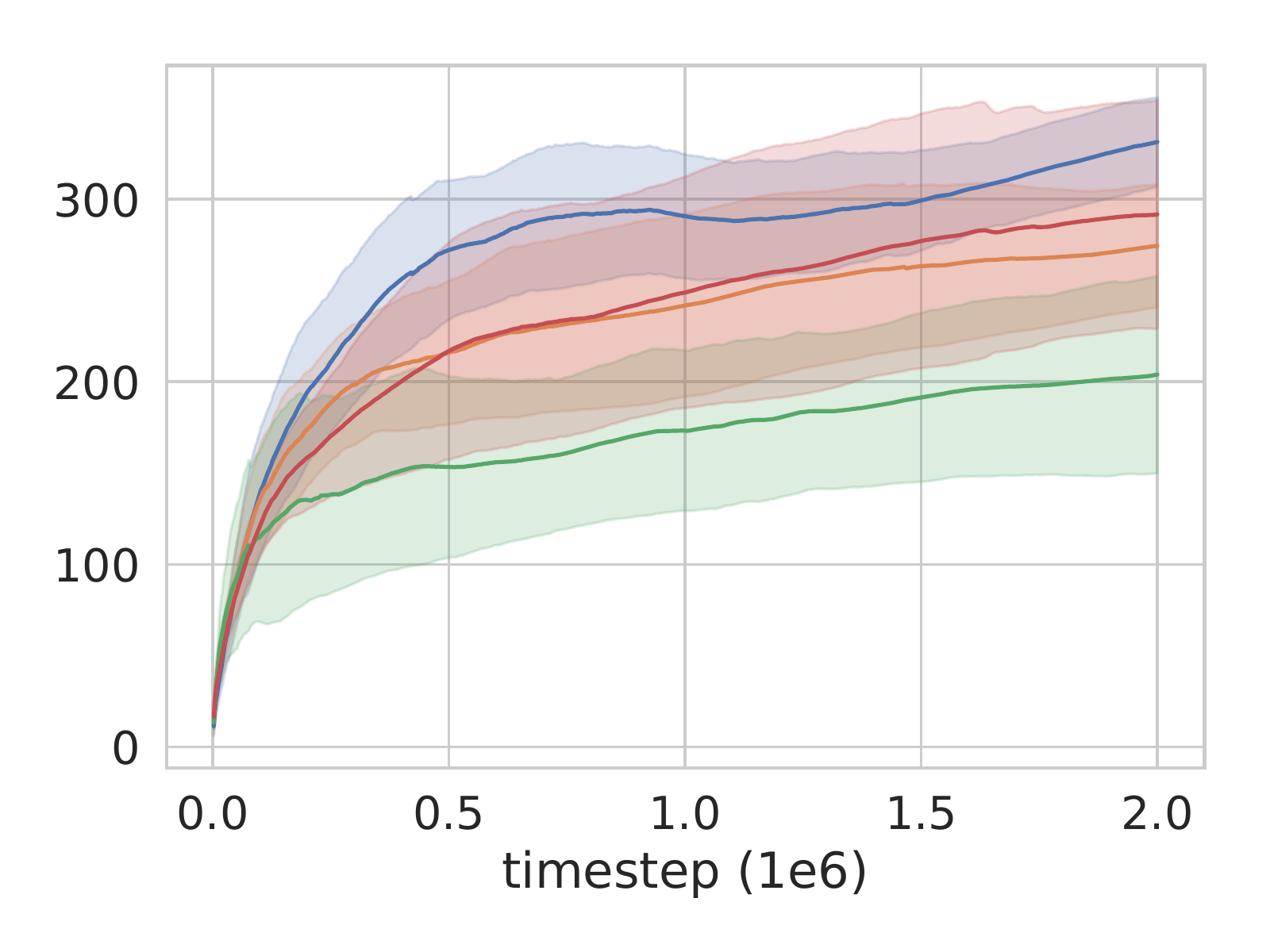}
    \caption{InvertedPendulum}
    \vspace{15pt}
  \end{subfigure}%
  \begin{subfigure}[b]{0.25\textwidth}
    \includegraphics[width=\textwidth]{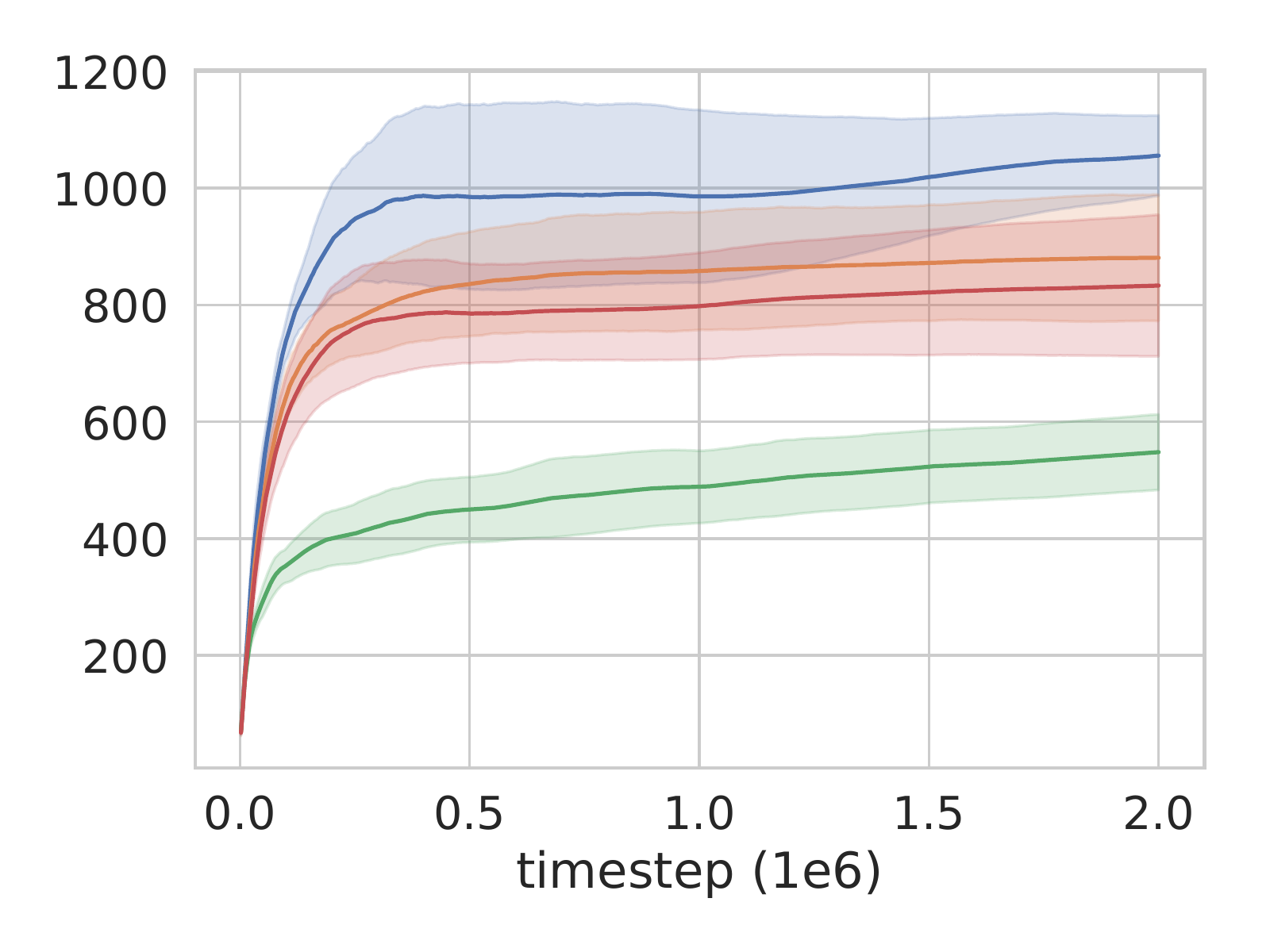}
    \caption{InvertedDoublePendulum}
    \vspace{15pt}
  \end{subfigure}%
  \begin{subfigure}[b]{0.25\textwidth}
    \includegraphics[width=\textwidth]{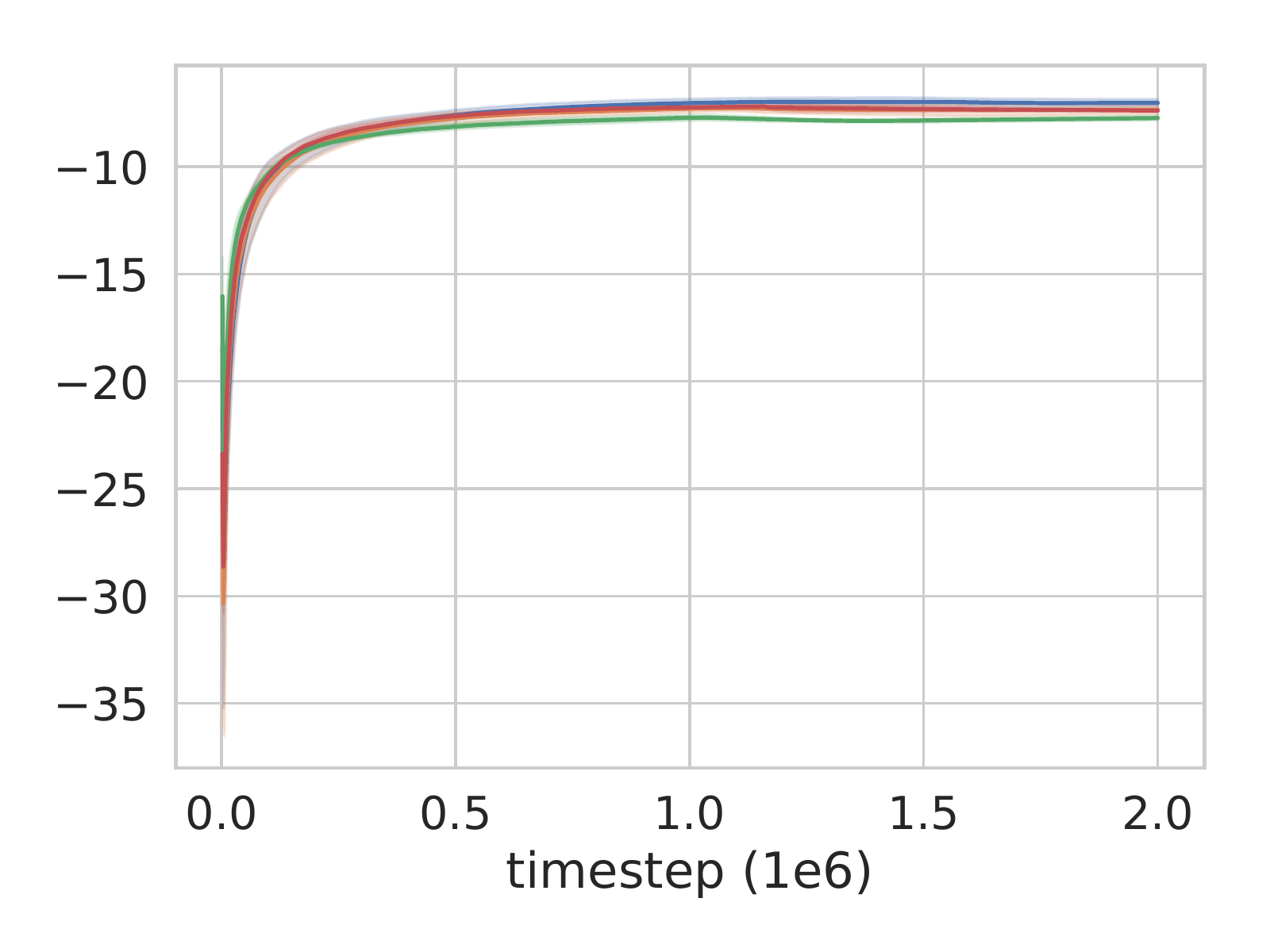}
    \caption{Reacher}
    \vspace{15pt}
  \end{subfigure}%
  
  \begin{subfigure}[b]{0.25\textwidth}
    \includegraphics[width=\textwidth]{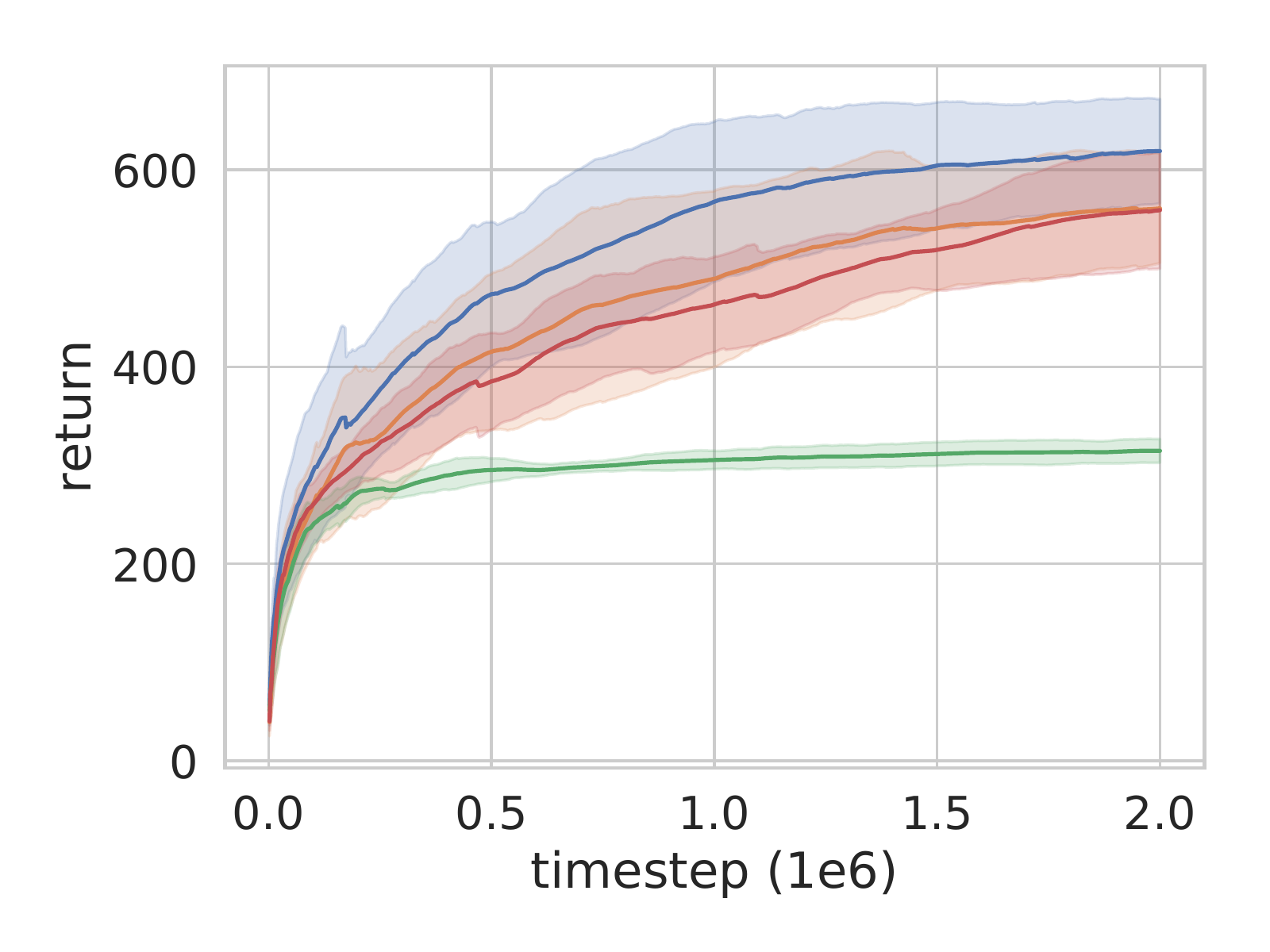}
    \caption{Hopper}
  \end{subfigure}%
  \begin{subfigure}[b]{0.25\textwidth}
    \includegraphics[width=\textwidth]{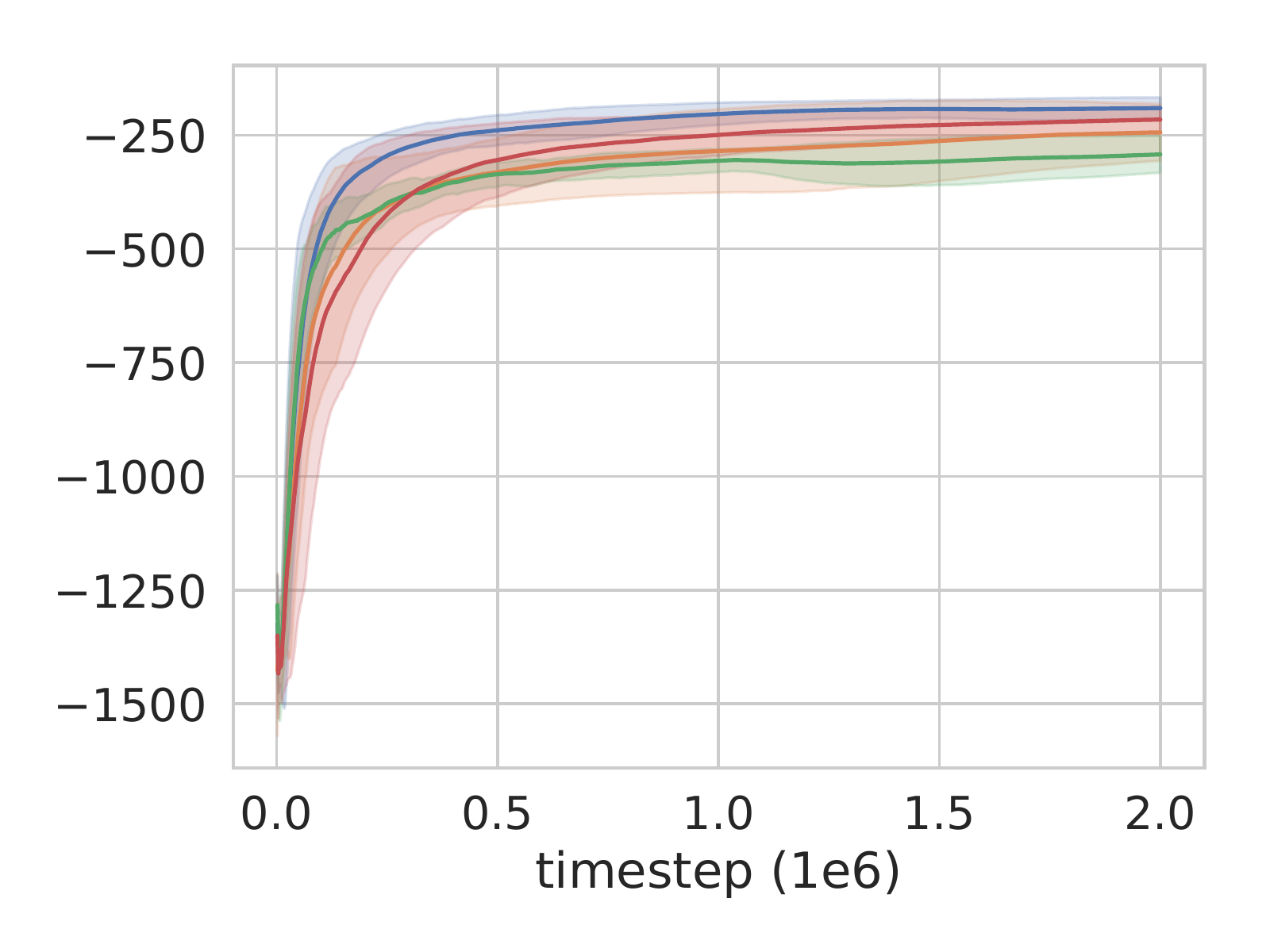}
    \caption{Pendulum}
  \end{subfigure}%
  \begin{subfigure}[b]{0.25\textwidth}
    \includegraphics[width=\textwidth]{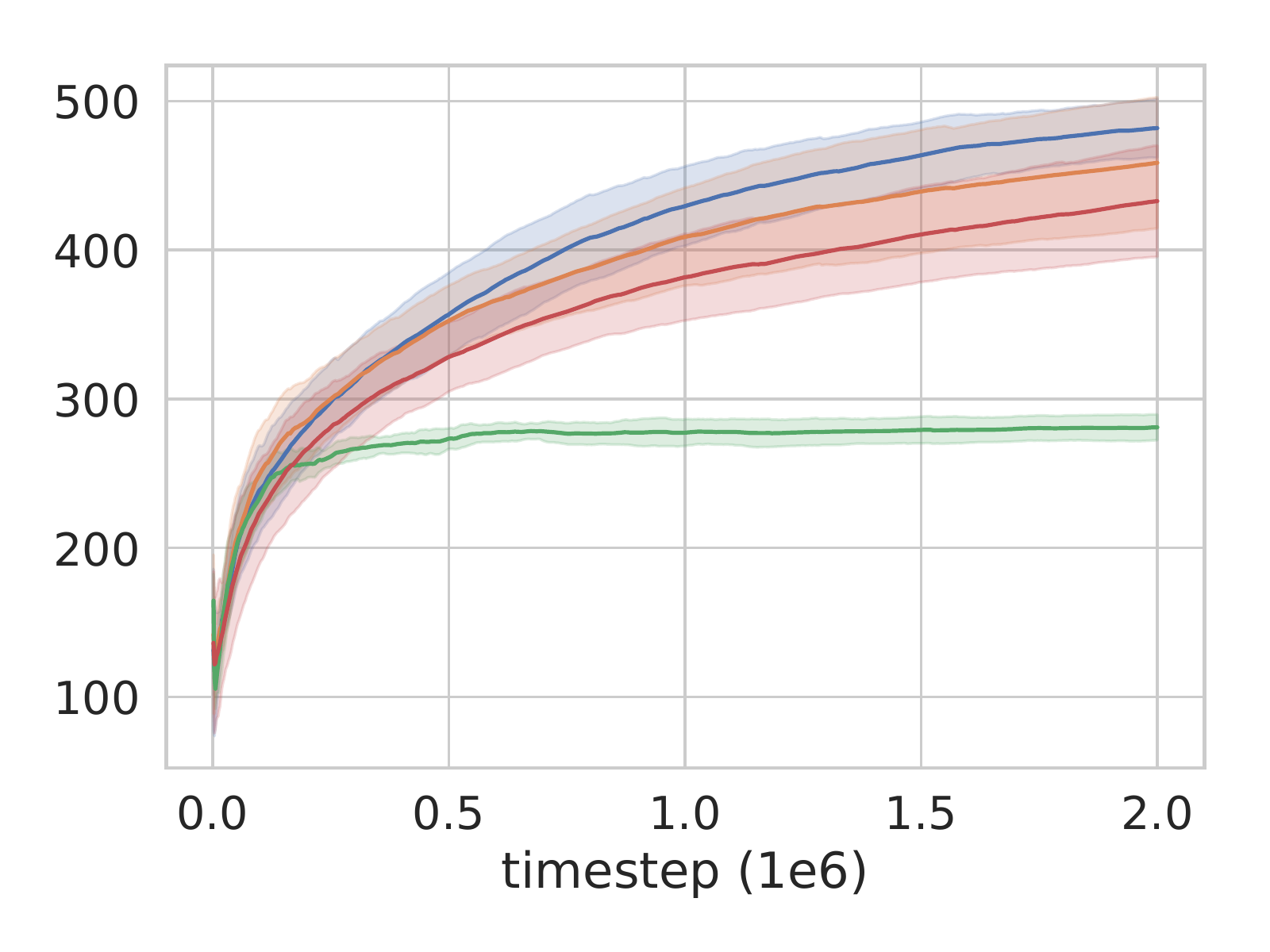}
    \caption{Humanoid}
  \end{subfigure}%
  \begin{subfigure}[b]{0.25\textwidth}
    \includegraphics[width=\textwidth]{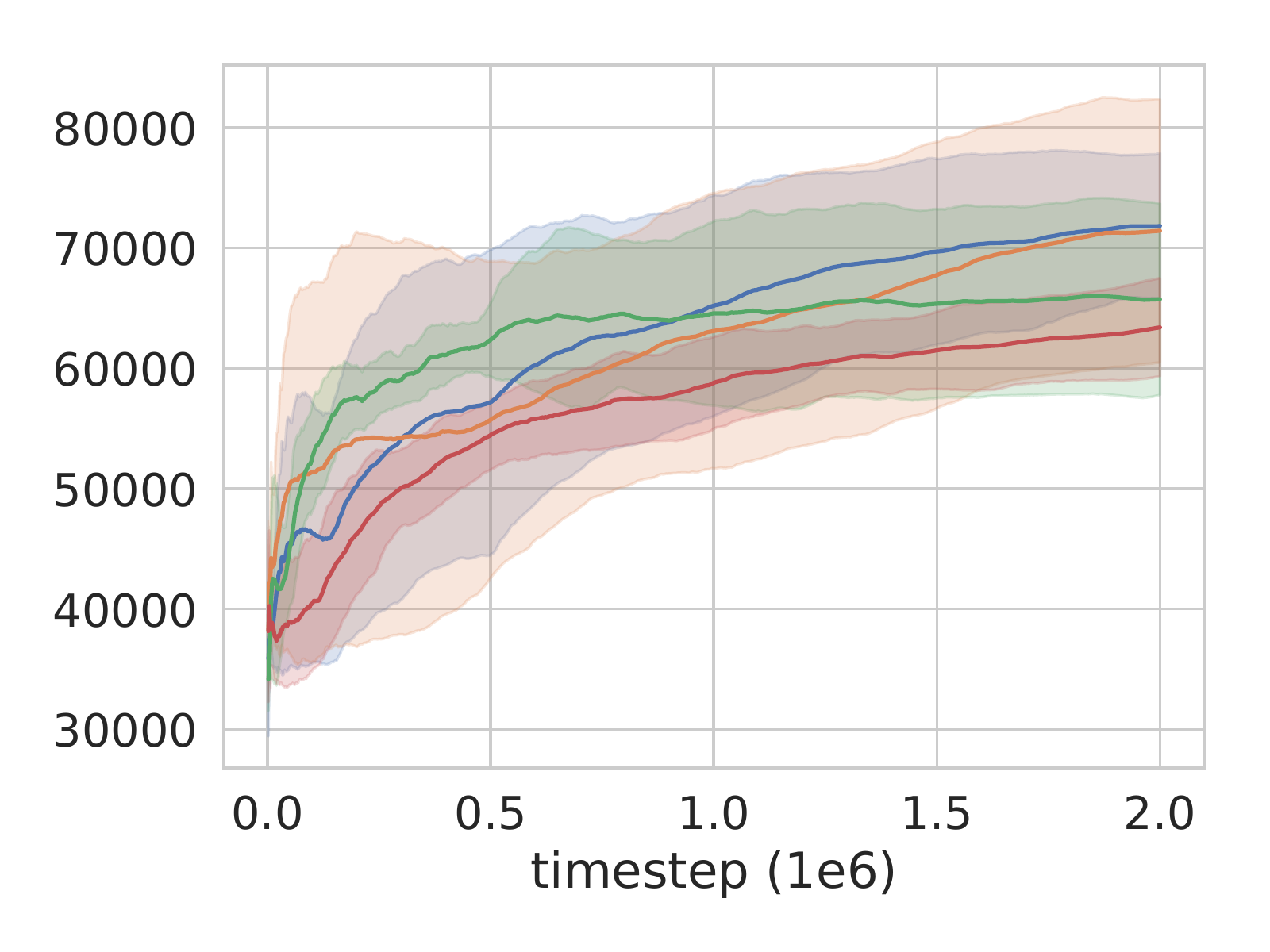}
    \caption{HumanoidStandup}
  \end{subfigure}%
  \caption{Performance comparison of ERO against baselines on 8 continuous control tasks. The shaded area represents mean $\pm$ standard deviation. ERO outperforms baselines on most of the continuous control tasks.}
  \label{label:performance}
\end{figure*}

\section{Experiments}
\label{label:sec_exp}
In this section, we conduct experiments to evaluate ERO. We mainly focus on the following questions: \textbf{Q1:} How effective and efficient is our ERO compared with the uniform replay strategy as well as the rule-based alternatives? \textbf{Q2:} What kind of transitions could be beneficial based on the learned replay policy?
\subsection{Experimental Setting}

Our experiments are conducted on the following continuous control tasks from OpenAI Gym\footnote{https://gym.openai.com/}: HalfCheetah-v2, InvertedDoublePendulum-v2, Hopper-v2, InvertedPendulum-v2, HumanoidStandup-v2, Reacher-v2, Humanoid-v2, Pendulum-v0~\cite{todorov2012mujoco,brockman2016openai}. We apply ERO to the benchmark off-policy algorithm DDPG~\cite{lillicrap2016continuous} for the evaluation purpose. Our ERO is compared against the following baselines:
\begin{itemize}
    \item \textbf{Vanilla-DDPG:} DDPG with uniform sampling.
    \item \textbf{PER-\emph{prop}:} Proportional prioritized experience replay~\cite{schaul2016prioritized}, which prioritizes transitions with high temporal difference errors with rules.
    \item \textbf{PER-\emph{rank}:} Rank-based prioritized experience replay~\cite{schaul2016prioritized}, a variant of PER-\emph{prop} which adopts a binary heap for the ranking purpose.
\end{itemize}
For a fair comparison, PER-\emph{prop}, PER-\emph{rank} and ERO are all implemented on the identical Vanilla-DDPG.

\paratitle{Implementation details} Our implementations are based on OpenAI DDPG baseline~\footnote{https://github.com/openai/baselines}. For Vanilla-DDPG, we follow all the settings described as in the original work~\cite{lillicrap2016continuous}. Specifically, $\tau=0.001$ is used for soft target updates, learning rates of $10^{-4}$ and $10^{-3}$ are adopted for actor and critic respectively, the Ornstein-Uhlenbeck noise with $\theta=0.15$ and $\sigma = 0.2$ is used for exploration, the mini-batch size is $64$, the replay buffer size is $10^6$, the number of rollout steps is $100$, and the number of training steps is $50$. For other hyperparameters and the network architecture, we use the default setting as in the OpenAI baseline. For the two prioritized experience replay methods PER-\emph{prop} and PER-\emph{rank}, we search the combinations of $\alpha$ and $\beta$ and report the best results. For our ERO, we empirically use three features for each transition: the reward of the transition, the temporal difference~(TD) error, and the current timestep. TD error is updated only when a transition is replayed. We implement the replay policy by using MLP with two hidden layers (64-64). The number of replay updating steps is set to $1$ with mini-batch size 64. Adam optimizer is used with a learning rate of $10^{-4}$. Our experiments are performed on a server with 24 Intel(R) Xeon(R) CPU E5-2650 v4 @ 2.2GHz processors and 4 GeForce GTX-1080 Ti 12 GB GPU.

\subsection{Performance Comparison}
The learning curves on 8 classical control tasks are shown in Figure~\ref{label:performance}. Each task is run for $5$ times with $2 \times 10^6$ timesteps using different random seeds, and the average return over episodes is reported. We make the following observations.

First, the proposed ERO consistently outperforms all the baselines on most of the continuous control tasks in terms of sample efficiency. On HalfCheetah, InvertedPendulum, and InvertedDoublePendulum, ERO performs clearly better than Vanilla-DDPG. On Hopper, Pendulum, and Humanmoid, faster learning speed with a higher return is observed. The intuition of our superior performance on various environments is that the replay policy of ERO is updated to replay the most suitable transitions during training. This also demonstrates that exploiting the improvement of the policy as a reward to update the replay policy is promising.

Second, rule-based prioritized replay strategies do not provide clear benefits to DDPG on the $8$ continuous control tasks, which can also be verified from the results in~\cite{novati2018remember}. Specifically, PER-\emph{prop} only provides a very slight improvement on $5$ out of $8$ tasks, and shows no clear improvement on the others. An interesting observation is that PER-\emph{rank} even worsens the performance on $4$ out of $8$ continuous control tasks. A possible explanation is that the rank-based prioritization samples transitions based on a power-law distribution. When the buffer size is large, the transitions near the tail are barely selected, which may lead to significant bias in the learning process\footnote{We have tuned the hyperparameters of the importance sampling trying to correct the potential bias, but we did not observe better performance.}. Recall that the two rule-based strategies prioritize the transitions with high temporal difference (TD) errors. A potential concern is that a transition with high TD error may substantially deviate from the current policy, hence introducing noise to the training, especially in the later training stages when the agent focuses on a preferred state space. This may partly explain why the two PER methods do not show significant improvement over Vanilla-DDPG. The results further suggest that rule-based replay strategies may be sensitive to different agents or environments. By contrary, the replay policy of ERO is updated based on the feedback from the environment, and consistent improvement is observed in various tasks.

\begin{figure}[t]
  \centering
  \includegraphics[width=0.45\textwidth]{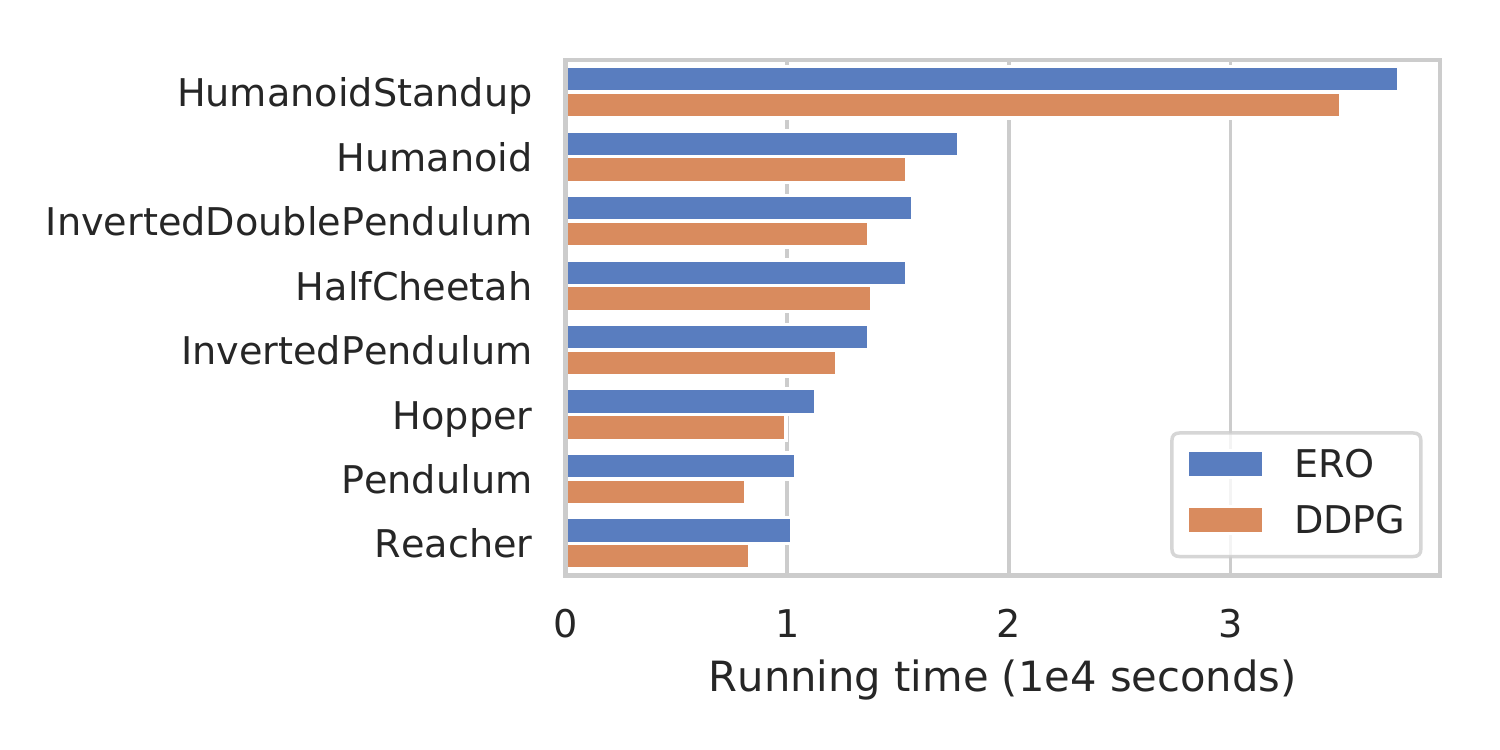}
  \caption{Comparison of running time in seconds for ERO and Vanilla-DDPG. ERO requires only slightly more running time compared with Vanilla-DDPG.}
  \label{label:time}
\end{figure}

\paratitle{Comparison of running time} Table~\ref{label:time} shows the average running time in seconds of ERO enhanced DDPG and Vanilla-DDPG on the 8 continuous control tasks for $2 \times 10^6$ timesteps. We observe that ERO requires slightly more running time than Vanilla-DDPG. It is expected because ERO requires additional computation for replay policy update. Note that the interactions with the environment usually dominate the cost of RL algorithms. In practice, the slightly more computational resources of ERO, which are often much cheaper than the interactions with the environment, are greatly outweighed by the sample efficiency it provides.

\begin{figure}[t]
  \centering
  \begin{subfigure}[b]{0.4\textwidth}
    \includegraphics[width=1.0\textwidth]{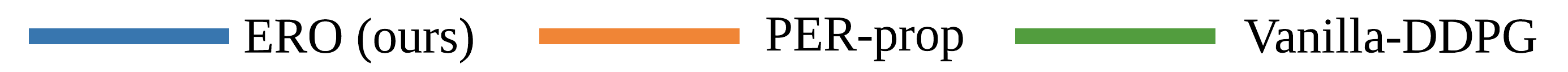}
  \end{subfigure}
  \vspace{5pt}
  
  \begin{subfigure}[b]{0.45\textwidth}
    \includegraphics[width=\textwidth]{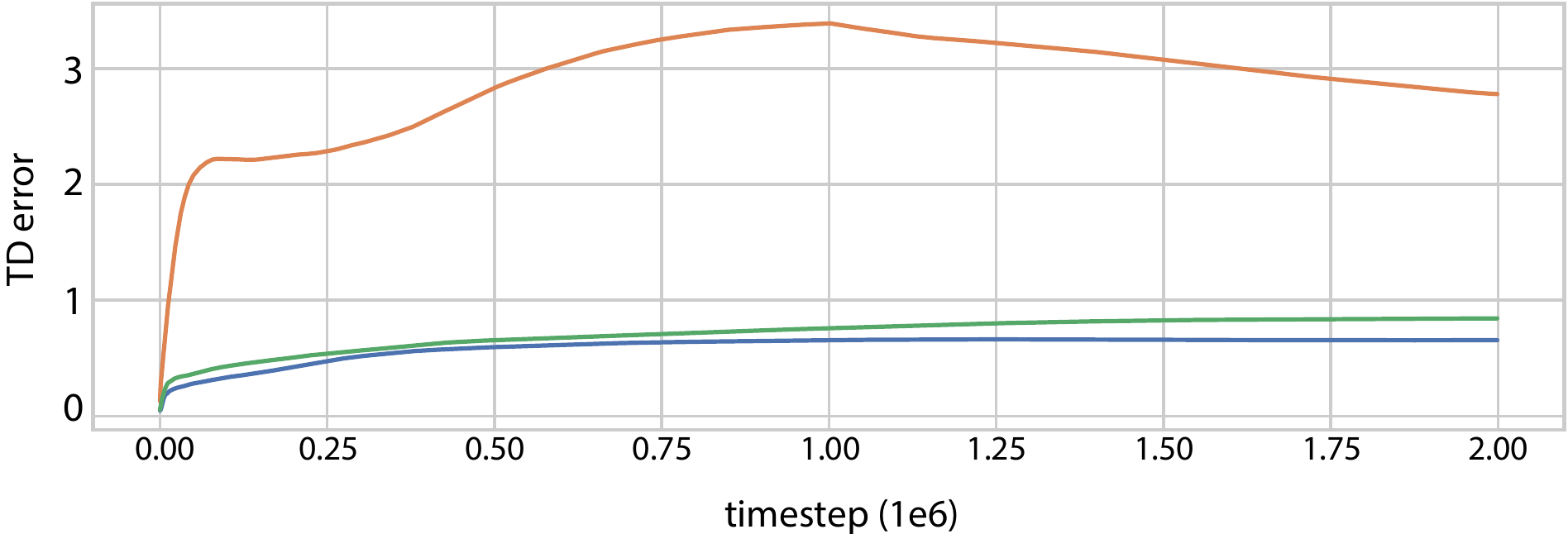}
  \end{subfigure}%
  
  \begin{subfigure}[b]{0.45\textwidth}
    \includegraphics[width=\textwidth]{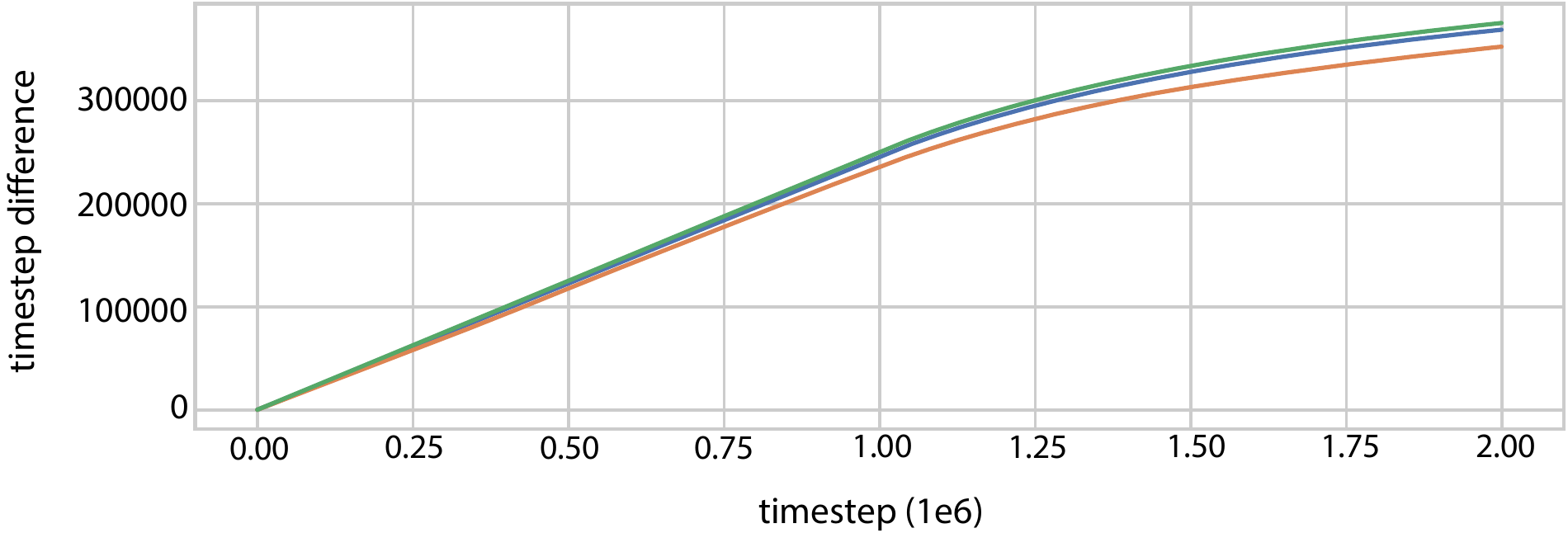}
  \end{subfigure}%
  
  \begin{subfigure}[b]{0.45\textwidth}
    \includegraphics[width=\textwidth]{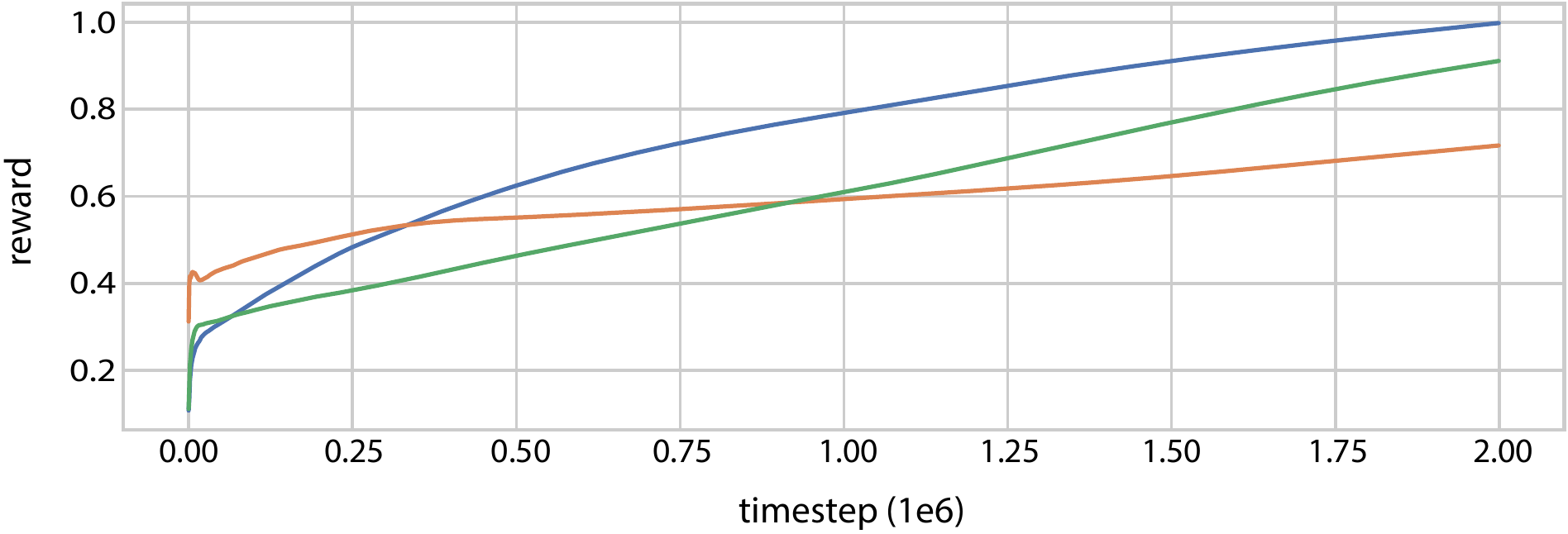}
  \end{subfigure}%
  
  \caption{The evolution of TD error~(top), timestep difference~(middle), and reward of the transition~(bottom) with respect to timestep. Timestep difference is the difference between the training timestep and the timestep at which the transition is obtained~(that is, low timestep difference means that the agent tends to use recent transitions for the updates). The average values over transitions are plotted for clearer visualization.}
  \label{label:transition}
\end{figure}

\subsection{Analysis of the Sampled Transitions}
For further understanding the learned replay policy, we record the transition features through the entire training process of one trial on HalfCheetah to analyze the characteristics of the sampled transitions. Figure~\ref{label:transition} plots the average values for the three features: temporal difference~(TD) error, timestep difference between the training timestep and the timestep at which the transition is obtained, and reward of the transition. We may empirically find some insights of the replay policy.

As expected, PER-\emph{prop} tends to sample transitions with high TD errors. Interestingly, the learned replay policy of ERO samples more transitions with low TD errors, which are even lower than Vanilla-DDPG. It contradicts to our belief that the transitions with low TD errors are well-known by the agent, which could cause catastrophic forgetting problem. It also contradicts to the central idea of prioritized replay which prioritizes ``unexpected" transitions. We hypothesize this task may not necessarily contain catastrophic states, and the transitions with slightly low TD errors may align better with the current policy and could be more suitable for training in this specific task. The unsatisfactory performance of PER methods also suggests that prioritizing transions with high TD errors may not help in this task. We believe more studies are needed to understand this aspect in the future work. We observe that both PER-\emph{prop} and ERO sample more recent transitions than Vanilla-DDPG. This is also expected for PER-\emph{prop}, because newly added transitions usually have higher TD errors. Although ERO tends to sample transitions with low TD errors, it also favors recent transitions. This suggests that recent transitions may be more helpful in this specific task. For the transition reward, all the three methods tend to sample transitions with higher rewards as timestep increases. It is reasonable since more transitions with higher rewards are stored into the buffer in the later training stages.

Overall, we find that it is nontrivial to heuristically use rules to define a proper replay strategy, which may depend on a lot of factors from both the algorithm and the environment. A simple rule may not be able to identify the most useful transitions and could be sensitive to different tasks or different reinforcement learning algorithms. For example, the prioritized experience replay shows no clear benefit to DDPG on the above continuous control tasks from OpenAI gym. A learning-based replay policy will be more desirable because it is optimized based on different tasks and algorithms.

\section{Discussion and Extension}
Replay learning can be regarded as a kind of meta-learning. There are several analogous studies in the context of RL: learning to generate data for exploration~\cite{xu2018learning}, learning intrinsic reward to maximize the extrinsic reward on the environment~\cite{zheng2018learning}, adapting discount factor~\cite{xu2018meta}. Different from these studies, we focus on how reinforcement learning agents can benefit from learning-based replay policy.

Our ERO can also be viewed as a teacher-student framework, where the replay policy~(teacher) provides past experiences to the agent~(student) for training. Our work has similarity to learning a teaching strategy in the context of supervised learning~\cite{fan2018learning,wu2018learning}. However, replay learning differs from teaching a supervised classifier. It is much more challenging due to the large and noisy replay memory, and the unstable learning signal. Our framework could be extended to teach the agent when to remember/forget experiences, or what buffer size to adopt.

ERO could potentially motivate the research of continual (lifelong) learning. Experience replay is shown to effectively transfer knowledge and mitigate forgetting~\cite{yin2017knowledge,isele2018selective}. While existing studies selectively store/replay experiences with rules, it is possible to extend ERO to a learning-based experience replay in the context of continual learning.

\section{Conclusion and Future Work}
In this paper, we identify the problem of experience replay learning for off-policy algorithms. We introduce a simple and general framework ERO, which efficiently replays the useful experiences to accelerate the learning process. We develop an instance of our framework by applying it to DDPG. Experimental results suggest that ERO consistently improves the sample efficiency compared with Vanilla-DDPG and rule-based strategies. While more studies are needed to understand experience replay, we believe our results empirically show that it is promising to learn experience replay for off-policy algorithms. The direct future work is to extend our framework to learn which transitions should be stored/removed from the buffer and how we can adjust the buffer size in different training stages. We are also interested in studying learning-based experience replay in the context of continual learning. We will investigate the effectiveness of more features, such as signals from the agent, to study which features are important. Finally, since our framework is general, we would like to test it on other off-policy algorithms.

\section*{Acknowledgements}
This work is, in part, supported by NSF (\#IIS-1718840 and \#IIS-1750074). The views and conclusions contained in this paper are those of the authors and should not be interpreted as representing any funding agencies.

\bibliographystyle{named}
\bibliography{reference}

\end{document}